\documentclass[letterpaper]{article}

\usepackage{aaai2027}
\nocopyright

\usepackage[hyphens]{url}
\usepackage{graphicx}
\usepackage{natbib}
\usepackage{caption}
\usepackage{booktabs}
\usepackage{xcolor}
\usepackage{amsmath}
\usepackage{amssymb}
\usepackage{amsthm}
\usepackage{multirow}
\usepackage{algorithm}
\usepackage{algorithmic}
\usepackage{tabularx}
\title{Beyond Asking: A Pipeline for Personalized Game\\ Generation that Reads Players from Behavior}

\author{
    Yifan Lu\textsuperscript{1},
    Xiaopeng Yuan\textsuperscript{2},
    Haohan Wang\textsuperscript{2}
}
\affiliations{
    \textsuperscript{1}Imperial College London\\
    \textsuperscript{2}University of Illinois Urbana-Champaign
}

\begin{document}

\maketitle

\begin{abstract}
Personalized game generation requires inferring a player's abilities and behavioral style from how they play. Large language models have made this inference more attainable than ever: an LLM can read a raw gameplay transcript and produce a fluent, plausible profile of the player. Plausible, however, is not verified, and verification is precisely what the field lacks: latent traits are unobservable; questionnaires provide noisy proxies and become circular when self-reports are used to validate behavior-based inference; and behavior itself is ambiguous without context—a player who never collects an item may not want it, or may never have had the chance.
We address both problems. First, we construct a synthetic player population whose traits are ground truth by construction: each trait is an explicit bot parameter, accepted only after controlled manipulation produces consistent, trait-specific behavioral change. Unlike prior parameter-recovery work that inverts a known decision model, our benchmark evaluates policy-agnostic inference from behavioral transcripts alone. Second, we introduce an opportunity-aware decision-moment representation that disentangles preference from the chance to express it; ablating it selectively degrades opportunity-dependent traits. On this benchmark, few-shot LLM inference outperforms embedding- and rule-based baselines on most traits, though feature-based supervised regressors remain stronger overall. Finally, we close the loop: inferred profiles drive difficulty adaptation, evaluated against ground-truth references and mismatched-profile controls, and an exploratory human study examines whether these findings transfer to real players.\looseness=-1

\end{abstract}

\section{Introduction}
Personalization is a long-standing goal of game content generation \cite{shaker2016procedural,yannakakis2018artificial}. If a system understands a player's abilities and behavioral style, it can generate levels with suitable difficulty and compatible gameplay characteristics \cite{yannakakis2011experience}. The development of large language models (LLMs) makes it feasible to infer latent behavioral tendencies directly from gameplay traces \cite{gallotta2024large,peters2024large}. An LLM can read a behavioral transcript and estimate a player's tendencies through natural-language reasoning, without requiring a separate classifier for each tendency dimension.

However, this approach faces an evaluation problem that precedes any specific modeling method: a player's latent behavioral tendencies are not directly observable, and their assessment has long relied on annotation via self-reports or external observers \cite{yannakakis2013player,goel2024labelfree,bunian2017modeling}. Whether the inference is performed by rules, supervised learning, or an LLM, it is therefore difficult to determine whether the inferred tendencies are correct. A common practice is to use questionnaires or self-reports as reference labels \cite{john1999big,zhu2025can}, but self-assessments are noisy and biased \cite{crowne1960new,podsakoff2003common,nisbett1977telling,yannakakis2011experience}. Moreover, when the motivation for behavioral inference is precisely to avoid reliance on self-reports, validating such inference against self-reports introduces a potential circularity. The consequence is that fluent player profiles can accumulate faster than the means to falsify any of them: an inference that is wrong in a systematic direction---capping one trait, inflating another---cannot be detected, let alone repaired. Prior work on procedural personas \cite{holmgard2014evolving,holmgard2019automated}, inverse reinforcement learning \cite{ng2000algorithms}, and cognitive modeling \cite{baker2009action,shergadwala2021can} shares a mature structure for exactly this situation: behavior is generated from known parameters, and an inference procedure is judged by whether it recovers them. Nevertheless, the emerging use of LLMs as behavior-to-tendency inference models still lacks an evaluation environment in which the ground truth is controlled, its validity is independently testable, and the boundaries of inference failure can be identified \cite{gallotta2024large}.

This paper presents an end-to-end personalization pipeline. It also provides a verifiable foundation for the component that has long remained hardest to evaluate: a synthetic diagnostic environment with controlled ground truth. Each behavioral tendency is implemented as an explicit generative parameter in the decision policy of a scripted agent, so the truth is determined by construction rather than by post hoc annotation. A known parameter, however, does not automatically constitute valid ground truth. We therefore introduce a ground-truth admission test: a parameter is accepted only if controlled manipulation produces monotone, dimension-specific behavioral change that exceeds seed-induced variation, measured on held-out behaviors that play no role in defining the parameter; failures are reported as rejected or non-identifiable. This converts recovery evaluation from a weakly supervised, difficult-to-falsify problem into a controlled and falsifiable one. Its scope is deliberately limited: the environment validates inference methods under controlled conditions rather than substituting for ground truth about real players, and transfer to real players is examined only preliminarily through an exploratory human study.

\begin{figure*}[t]
    \centering
    \includegraphics[width=\textwidth]{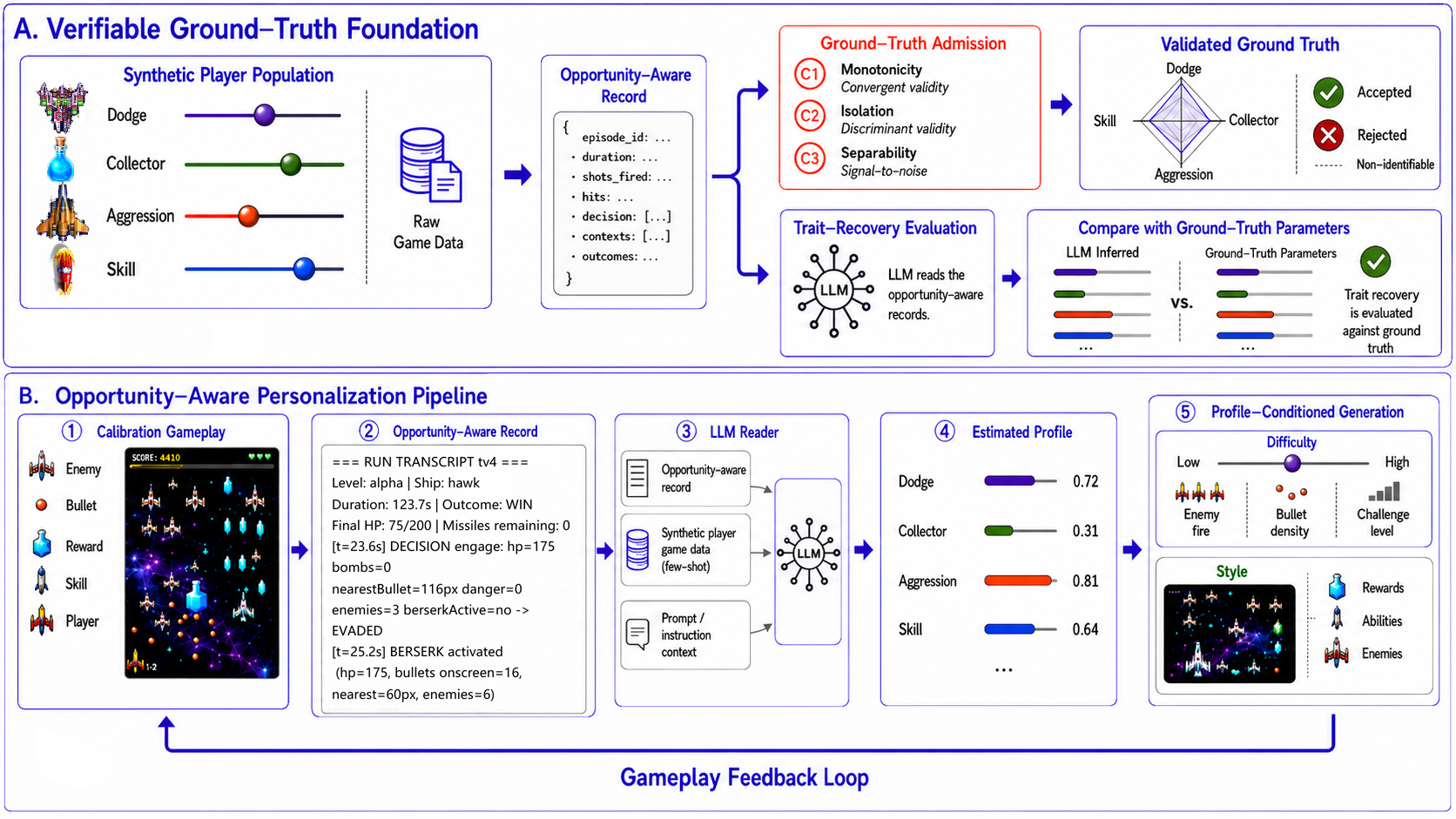}
    \caption{Overview. (A) Synthetic players with parameterized traits
produce opportunity-aware records; parameters enter the benchmark
only after passing the admission test, and trait recovery is scored
against them. (B) The same record schema drives the deployed
pipeline: calibration play, LLM reading, and profile-conditioned
generation.}
    \label{fig:pipeline}
\end{figure*}

We instantiate the diagnostic environment in a vertically scrolling
shooter. The genre is chosen for measurability: traits are observable
only where the situation affords their expression
\cite{tett2003personality}, and this genre supplies such
opportunities densely, as discrete decision moments at a fixed rate,
each with a choice set the engine already enumerates. At each
decision point the record keeps the game state, the set of feasible
choices, and the action actually selected, so that behavioral
tendency is separated from behavioral opportunity; an LLM reads these
records to infer four trait dimensions, and a rule layer maps the
inferred profile to generated levels. Nothing in the pipeline is specific to the genre: it transfers to any
game whose play can be translated into the same record schema, and
what changes across games is the translation layer, not the reading
or the validation around it. Our contributions are:
\begin{itemize}\setlength{\itemsep}{0pt}
\item A \textbf{controlled synthetic player population} whose four
  trait parameters are set by construction, providing the reference
  ground truth that LLM-based player modeling has so far lacked.
\item An \textbf{opportunity-aware record} that keeps declined game action choices on file beside taken ones; ablating the choice sets
  degrades exactly the traits that depend on opportunity.\looseness=-1

\item An \textbf{LLM player model} that reads the opportunity-aware record and infers
  a four-dimensional trait profile from a fixed example library
  collected once, with no gradient updates and no access to the
  generating policy.
\item A \textbf{closed-loop personalization experiment}: the
  LLM-inferred profile conditions a generator that assembles a level
  matched to that player's traits, and in a pilot with 12
  participants, levels generated from the behavioral profile score
  higher than levels generated from a questionnaire.
\end{itemize}

\section{Related Work}
\paragraph{Inferring latent traits, and what serves as truth.}
LLMs have been used to infer personality and other latent variables
from text or behavioral records
\cite{zhu2025evaluating,zhu2025can}, but the ground truth in these
studies comes almost entirely from self-report scales, the very
criterion this paper tries to avoid; psychometric surveys point out
the resulting validity problems \cite{ye2026large}. Reading ability
from behavior rather than from asking has a precedent in
evidence-centered design and stealth assessment
\cite{mislevy2003ecd,shute2013stealth}, where task performance is
treated as evidence for a latent competence model. We do not measure
the personality of an LLM itself; we measure its ability as a reader
where the truth is controlled.

\paragraph{Synthetic players are not new, but their role here is.}
The procedural personas tradition
\cite{holmgard2014evolving,holmgard2015monte,holmgard2019automated}
implements play styles as utility weights and sends evolved personas
to play levels in place of human testers, for the purpose of content
evaluation. Closest to our work is \citet{shergadwala2021can}, which
implements behavior tendencies as parameters of a cognitive decision
model and recovers them by inverse Bayesian inference. We differ on
three points. Personas serve content evaluation and, to our
knowledge, have not been used as a ground-truth benchmark for
trait-recovery systems. Both lines treat their parameters as ground
truth by definition, whereas we admit a parameter only after it
passes a behavioral consistency test. And their inference holds the
generative model itself, a form of same-model inversion, whereas our
readers have no access to the generating policy. Generating behavior
from known parameters and testing recovery is otherwise a standard
structure \cite{ng2000algorithms,baker2009action,talts2020validating};
what is new is its organization into a protocol where the ground
truth must first be admitted and failures can be localized.

\paragraph{Player modeling and content generation.}
Both have been studied for decades
\cite{yannakakis2018artificial,shaker2016procedural,bunian2017modeling},
and dynamic difficulty adjustment has long closed the loop between
them \cite{hunicke2005dda,mortazavi2024dda,lopes2025closing}. What
the loop has never had is a way to check its own first step: recent
surveys note that player modeling is nearly absent from LLM research
\cite{gallotta2024large}, and the LLM work that does touch games goes
straight to generation
\cite{sudhakaran2023mariogpt,todd2023llm}. Our generation side is
deliberately simple, because the contribution is not a new generator
but a reading stage whose output can be validated before it drives
one.

\section{Problem and Pipeline}
The setup is a recovery problem. A synthetic player whose traits we
set plays a level; the session is written down in a form that keeps
both what the player did and what was available to do; a LLM reader tries
to recover the traits from that record alone, and because we set the
traits, its estimate can be checked. The rest of this section states
this precisely.\looseness=-1

We characterize a player by a latent vector of behavioral tendencies.
At each decision moment $t$ during play, the player faces a state
$s_t$ and a set of currently feasible choices $A_t$, and takes an
action $a_t$. Behavior is generated by an unknown policy:
\[a_t \sim \pi(\,\cdot \mid s_t,\ A_t;\ \theta\,), \qquad \theta \in [0,1]^4.\]
The central point is that observed behavior is determined jointly by
tendency and opportunity: the same $\theta$ produces different
behavior under different $A_t$. If a record keeps only what was done
and discards what could have been done, tendency and opportunity
become inseparable. A session is therefore recorded as a sequence of
opportunity-aware decision moments:
\[\tau = \{(s_t,\ A_t,\ a_t)\}_{t=1}^{T}, \qquad \hat\theta = f(\tau),\]
where $f$ is the reader under evaluation, instantiated here mainly as
a large language model. Recovery quality is measured by
per-dimension rank correlation $\rho(\hat\theta_k,\theta_k)$ and
per-dimension error. Downstream, $\hat\theta$ maps through a fixed
function to a target difficulty $D^*=g(\hat\theta)$, and the
generator brings the assembled level close to that target.

The pipeline runs calibration play, opportunity-aware recording,
trait reading, difficulty mapping, and profile-conditioned
generation, with replay feeding back into the record
(Figure~\ref{fig:pipeline}B). Each stage is checked against
controlled ground truth, which is where the four research questions
come from.

\section{Method}
The four parts answer requirements that follow from one another.
Traits must be explicit parameters entering separable channels, or no
reader could tell them apart (Synthetic Players). A parameter is not
a valid trait until it survives a test on behavior it did not define
(Ground-Truth Admission). A reader can only recover what the record
preserves, and behavior without its choice set is ambiguous (Records
and Reading). And an inferred profile is only useful if it changes
the level the player receives (Profile-Conditioned Adaptation).
\subsection{Synthetic Players}
 
We need players whose traits are caused rather than labeled: behavior
must follow from a parameter we set, not from an annotation assigned
afterward. Each synthetic player is a real-time decision maker in a
bullet-hell shooter running at sixty frames per second, with the
decision module decoupled from the rendering engine and controlled by
a single random seed, so that the trait parameters and the seed
reproduce the complete trajectory frame by frame.\looseness=-1

The policy has three components, kept separate so that the traits do
not interfere: if two traits shared a channel, no reader could
separate them, and the admission test below would reject both.
Objective arbitration decides when to leave avoidance and pursue an
objective, with pursuit intent resampled every two seconds from the
preference parameters; this stochastic expression proved superior to
persistent lock-on, which decouples the parameter from behavior. The
movement layer scores candidate positions by a danger field and
vetoes directions failing a time-to-collision gate. The action layer
evaluates discrete abilities at fixed intervals, first judging
whether an opportunity exists. Full definitions are in the appendix.
 
\paragraph{Trait parameterization.}
Each of the four trait parameters $\theta_k \in [0,1]$ enters exactly
one channel: Dodge the movement layer, Collector and Aggression the
arbitration module, Skill the action layer. The preference dimensions
enter through an intent gate, pursuing when $u_t < \theta_k$ with
$u_t \sim U(0,1)$ resampled every two seconds, so the expected share
of pursuit time equals $\theta_k$; both also shrink the safety margin
required before pursuit, $T_{\text{safe}}(\theta_k) = T_0(1 -
0.5\,\theta_k)$, and Aggression additionally controls press depth
beneath enemies. The Dodge dimension linearly scales four
movement-layer coefficients (reaction interval, lookahead, perception
radius, motor noise), of which the reaction interval is the most
consequential:\looseness=-1

\begin{equation}
\Delta t(\theta_{\text{dodge}}) = 420 - 340\,\theta_{\text{dodge}}
\ \text{ms}.
\label{eq:reaction}
\end{equation}
The endpoints bracket the human range rather than imitate it:
420\,ms is about twice the mean simple visual reaction time of a
1469-adult sample \cite{woods2015factors}, and 80\,ms is below any
human latency, so the sweep spans impaired to superhuman rather than
compressing six levels into the human band. Between two decisions the
agent executes its stale target blindly, so a slow reactor walks into
bullets that appear inside the decision gap. The Skill dimension is
the most direct channel, a conditional use probability
$P(\text{use} \mid \text{opportunity}) = \theta_{\text{skill}}$,
where the thresholds defining an opportunity are fixed and do not
vary with the parameter.
 
\subsection{Ground-Truth Admission}
 
Writing a parameter into code does not make it ground truth: it may
fail to produce the intended behavioral effect, or may be entangled
with other parameters. We therefore subject every parameter to an
admission test before it enters the benchmark. Let $m_k$ denote the
signature behavioral metric of dimension $k$, chosen to be
\emph{held out}: it plays no role in defining the parameter itself.
Each parameter is swept over $L=6$ levels
($\theta_k \in \{0, 0.2, \dots, 1.0\}$) with five random seeds per
level ($n=30$ runs per dimension), and admitted only if three
conditions hold together, instantiating construct validation
\cite{cronbach1955construct} in a setting where the construct is
known by construction. \textbf{C1, monotonicity} (convergent
validity): the parameter correlates monotonically with $m_k$ across
levels, at Spearman $\rho \ge 0.6$ with no significant non-monotone
segment. \textbf{C2, isolation} (discriminant validity
\cite{campbell1959convergent}): varying the parameter induces no
systematic drift in the signature metrics of the other dimensions.
\textbf{C3, separability}: between-level differences exceed
seed-induced variance, quantified by an effect-size ratio and
adjacent-level discriminability. Exact statistics and thresholds for
C2 and C3 are in the appendix. Configurations that fail do not enter
the benchmark and are reported as rejected or non-identifiable; we
froze all metrics, directions, and thresholds before observing any
model response.
 
\subsection{Opportunity-Aware Records and Trait Reading}
 
We serialize each session into a plain-text record whose schema is
the same for every level, so that content regenerated per player
still yields the same kind of readable object. The design principle
is to keep opportunities and choices on file together: recording only
what the player \emph{did} is insufficient for judging what the
player \emph{wanted}, since the count of opportunities is the
denominator against which willingness is judged. What the record does
\emph{not} contain is the outcome an unselected action would have
produced; the representation is choice-set-aware, not counterfactual.
 
A record contains decision-moment lines with declined opportunities
alongside taken ones, combat events, windowed summaries of
positioning and opportunity density, and a closing tally with the
fixed game constants needed as denominators. An excerpt appears in
the appendix, including a collection opportunity the agent declines,
on file precisely so that \emph{did not} can be told apart from
\emph{could not}. Serialization follows two disciplines: no trait or
metric name ever appears in a record, enforced by a blacklist check;
and the wording follows the empirical distribution of behavior rather
than the visual layout of the screen.
 
Because personalized generation means that no two players see the
same content, a reader fitted to one level's statistics would need
refitting for every level it produces, and no labels exist for
content that does not exist until it is generated. The reader is
therefore prompted rather than trained. It operates in a single
few-shot setting \cite{brown2020language}: the system prompt contains
the game rules, behavioral definitions of the four dimensions, and
scoring guidance grounded in quantities countable from the record;
the context adds a table of roughly one hundred rows mapping
behavioral metrics to verified parameters, and the two most similar
sessions retrieved by behavioral features, each with its verified
profile \cite{liu2022makes}. The example library and the 90-session
test set are disjoint at the run level, with the query session and
any run sharing its trait configuration excluded from
retrieval.\footnote{The reader does not update model parameters but
does consume labeled data in context, so we avoid describing the
setting as zero-training.} Each session is read three times and the
scores averaged, with the session as the unit of inference.
 
\subsection{Profile-Conditioned Adaptation}
 
An inferred profile is worth nothing unless it changes what the
player receives, and unless that change can be checked by something
the generator does not control. Dodge and Aggression drive
difficulty: a response surface fitted on measured data maps them to a
target difficulty, which is converted into a pair of enemy-fire
settings and applied uniformly at level assembly, deterministically,
so the same profile locks to the same difficulty across repeated
generations. Collector, Skill, and a flanking facet of Aggression are
expressed through a content style layer that adjusts reward
placement, ability-opportunity cadence, and enemy approach angles;
the templates are in the appendix.
 
To separate \emph{matching} a player from merely \emph{easing} the
game, we define a difficulty coefficient
\begin{equation}
D_L = \sigma\!\Big(\textstyle\sum_i w_i
  \ln \frac{x_i}{x_i^{\mathrm{ref}}}\Big) \in [0,1],
\label{eq:difficulty}
\end{equation}
the sigmoid of a weighted sum of log ratios of level content features
$x_i$ against a standard calibration level, anchored so that the
standard level sits at $D_L = 0.5$. The weights $w_i$ are fixed a
priori from measured feature elasticities, fire pressure being the
dominant lever, rather than learned, so that $D_L$ is an
interpretable content-pressure coefficient rather than a fitted
success predictor. On this coefficient a novice receives a level with
$D^* \approx 0.13$ and an expert receives $D^* \approx 0.57$:
matching goes in both directions, while easing goes in only one.
$D_L$ controls generation only; downstream evaluation uses
independent behavioral outcomes, so the system is not measured with
its own ruler.\looseness=-1

Those outcomes are scored on a flow-matching scale
\cite{csikszentmihalyi1990flow,chen2007flow} where lower is better: a
run whose hits per minute falls inside a band frozen in advance from
mid-level agents scores zero, deviation outside the band scores up to
one, and a run ending in death scores above one, ranked by how early
it ended. The jump at one is deliberate, since hits per minute cannot
distinguish a death from a short clean run. The band is a fixed
experiential target, so adaptation is judged by whether it brings
agents of any level into it, not by whether each level meets a
level-specific criterion. The piecewise definition is in the
appendix.

\section{Experiments}
\subsection{Setup}
\noindent\textbf{Data.} Ground-truth admission uses 30 runs per
dimension (six levels, five seeds each). Recovery uses 210 labeled
training sessions and a frozen test set of 90. The identifiability
analysis uses 300 randomly parameterized sessions, the
opportunity-context ablation 300 paired sessions, and the diagnostic
probe 30 sessions per condition.

\noindent \textbf{LLMs and Baselines.} We evaluate two LLM readers, GPT-5.6-sol \cite{openai2026gpt56} and Qwen3.5-122B-A10B-think \cite{qwen2026qwen35}, and compare them with rule-based, supervised, retrieval-based, and statistical baselines. Rule uses handcrafted behavioral signatures. FeatReg applies RBF kernel ridge regression
\cite{hoerl1970ridge,scholkopf2002learning} over behavioral features
(a linear-ridge variant scores slightly lower); EmbReg applies kernel ridge over transcript embeddings from the
OpenAI \texttt{text-embedding-3-small}
model,\footnote{\url{https://platform.openai.com/docs/guides/embeddings}}
an approach reported to work well for text-based trait inference
elsewhere \cite{maharjan2025psychometric}, which makes its performance on our records (RQ2) informative rather than a weak-baseline artifact. BPM 1-NN matches each session to its
nearest labeled behavioral prototype \cite{cover1967nearest}. The
soft and hard GMM baselines \cite{dempster1977em} estimate traits
using eight behavioral clusters.

\noindent\textbf{Implementation.} Prompt construction, retrieval
exclusion, and three-read averaging follow the Method section. We
report per-dimension Spearman $\rho$ and MAE with macro averages.

\subsection{RQ1: When Does a Parameter Qualify as Ground Truth}
\begin{figure}[htbp]
  \centering
  \includegraphics[width=\columnwidth]{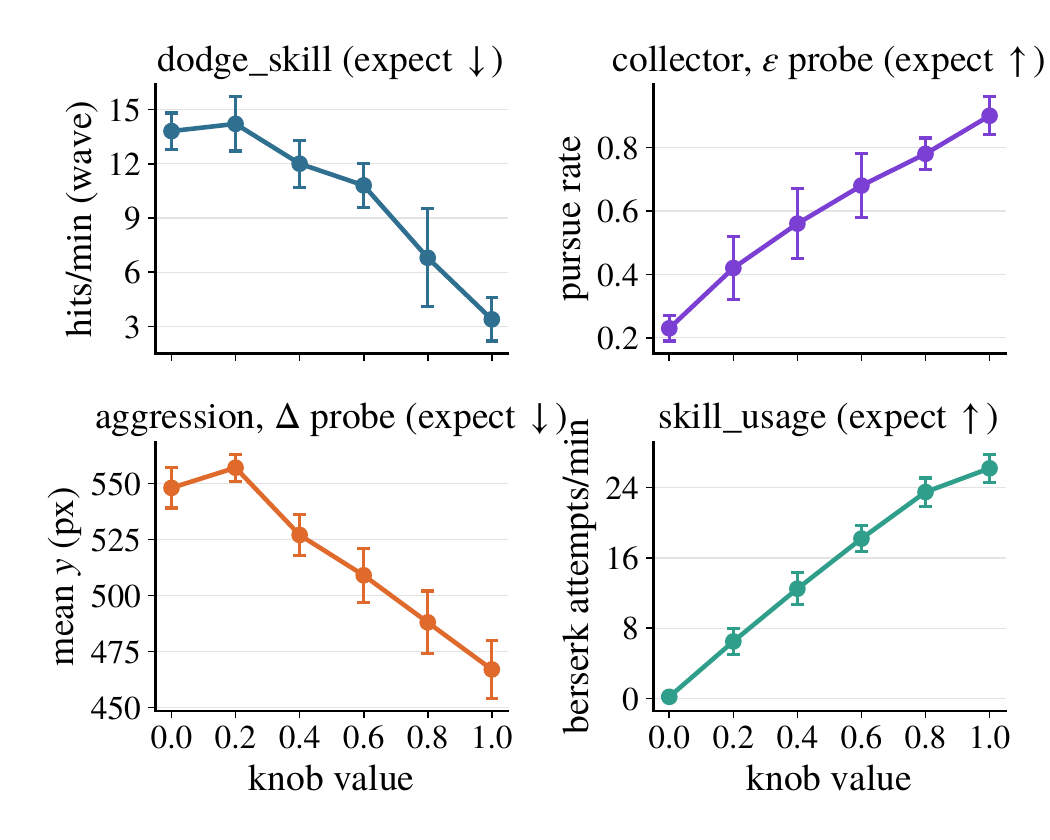}
  \caption{Trait-parameter monotonicity validation: each parameter
  produces the expected monotone trend in its held-out behavioral
  probe (mean $\pm$ std over 5 seeds per level, 6 levels each).\looseness=-1}
  \label{fig:knob-validation}
\end{figure}

Before asking whether a reader can recover a trait, we have to ask
whether the trait is there to be recovered. A parameter that produces
no consistent behavioral signature is not ground truth, whatever the
code says.
All four dimensions pass the admission threshold with monotone
responses (Figure~\ref{fig:knob-validation}), but their
distinguishable resolutions differ markedly. This is the point of the
admission test: it exposes weak constructs rather than hiding them.\looseness=-1

We further analyze multidimensional identifiability on 300 sessions
with random parameters, defining the behavioral-twin collision ratio
\begin{equation}
r_k = \frac{\mathbb{E}\big[\,|\Delta\theta_k| \mid
  \text{behavioral nearest-neighbor pairs}\,\big]}
  {\mathbb{E}\big[\,|\Delta\theta_k| \mid \text{random pairs}\,\big]},
\label{eq:collision}
\end{equation}
where a ratio near 1 means that individuals with the same behavior
are unconstrained in that dimension, i.e., observationally
equivalent. Skill is easiest to identify ($r=.51$) and Collector
hardest ($r=.82$): two agents with nearly identical behavior can hold
collector parameters far apart. The feature signal of Aggression is
in fact ample---positional evidence shows the strongest effect among
all metrics---so its recovery difficulty stems mostly from the reading stage
rather than from a lack of signal in the environment.

That a linear regressor recovers these parameters from a few hundred
labeled sessions is expected rather than surprising: the admission
test selects precisely for parameters whose behavioral effect is
consistent, monotone, and separable. A clean feature-to-parameter
mapping is what an admitted parameter means; parameters without one
were rejected before the benchmark was formed.

\subsection{RQ2: Can External Readers Recover the Parameters}
With ground truth admitted, the question becomes what can recover it
from behavior alone, and how much of any shortfall belongs to the
reader rather than to the record.
On a frozen test set of 90 sessions (a preregistered split), all
methods read the same complete records and are scored on the same
partition; Table~\ref{tab:recovery} reports Spearman $\rho$ per
dimension.

\begin{table}[htbp]
\centering
\small
\setlength{\tabcolsep}{4.5pt}
\begin{tabular}{lccccc}
\toprule
Method & Dodge & Coll. & Aggr. & Skill & Macro \\
\midrule
Rule            & .67 & \underline{.50} & .45 & .89 & .63 \\
FeatReg         & \textbf{.77} & \textbf{.56} & \textbf{.83} & \textbf{.91} & \textbf{.77} \\
EmbReg          & .65 & $-.04$ & .23 & .67 & .38 \\
BPM 1-NN        & .53 & .32 & .58 & .84 & .56 \\
GMM (soft, $K{=}8$) & .36 & .38 & .62 & .76 & .53 \\
GMM (hard)      & .32 & .38 & .56 & .75 & .50 \\
\midrule
GPT-5.6-sol & \underline{.69} & .43 & \underline{.79} & \underline{.90} & \underline{.70} \\
Qwen3.5-122B & .62 & .49 & .69 & .82 & .65 \\
\bottomrule
\end{tabular}
\caption{Recovery on the frozen test partition (Spearman $\rho$).
Best in \textbf{bold}, second best \underline{underlined}. Supervised
baselines are trained on 210 labeled sessions drawn from the same
distribution as the test set; the LLM rows are our few-shot protocol
instantiated with the indicated backbone.}
\label{tab:recovery}
\end{table}
Three observations follow. First, feature regression is best read as
an instrument rather than a competitor. It answers a different
question: given labels drawn from the same distribution as the test
set, how much trait information do these records contain at all? Its
macro $\rho$ of .77 answers that they contain a great deal, which is
what makes every other reader's shortfall interpretable; had it
reached only .40, the failure would lie in the representation and no
reader could do better. The supervised baselines are trained on 210
labeled sessions and evaluated on the held-out 90, both partitions
drawn from the same distribution, so this instrument presumes a
labeled corpus in-distribution with the test set, a condition that
cannot hold for real players whose true traits are unobservable.

Second, the LLM reader with GPT-5.6-sol attains $.70$ macro, second
overall, while consuming only a fixed, offline example library and no
gradient updates; the same protocol on Qwen3.5-122B-A10B-think
reaches $.65$, above the rule baseline, so the protocol is not tied
to a single backbone. Third, Collector is the hardest dimension for
every method, consistent with its collision ratio in
Eq.~\ref{eq:collision}: the difficulty is in the environment's
expressiveness, not in any particular reader.

The embedding regression reads the same transcript text but pools it
into a single fixed-length vector. The opportunity--choice structure
that distinguishes a declined pickup from an absent one is a local,
countable relation between specific lines, and mean-pooling averages
it away; the collapse is sharpest on Collector ($\rho\approx0$),
whose signal lives almost entirely in that relation, while dimensions
whose signal survives as global statistics (Dodge, Skill) retain
moderate correlation.

\paragraph{Where the reader loses, and why.}
The gap to feature regression is uneven across dimensions, and its
shape identifies its cause. On Skill the two are level (.91 versus
.90): the evidence is a countable frequency that survives
serialization intact. Aggression is nearly level as well (.83 versus .79). The gap on Dodge (.77 versus .69) tracks quantization:
grazing distance and reaction latency are continuous quantities that
the transcript reports in coarse categories, which a regressor reads
directly from the numeric feature. The largest gap is on Collector
(.56 versus .43), and it is environmental rather than architectural:
on the diagnostic probe level, where collection opportunities
actually arise, the same reader attains $\rho=.92$
(Table~\ref{tab:probe}), matching the handcrafted signature and
exceeding linear ridge regression. What the standard level denies
the reader is opportunity, not capability.

\subsection{RQ3: Does Failure Lie in the Environment}
\label{sec:rq3}

A recovery score that falls short says nothing about where to look.
Two explanations are always available: the record may not carry the
evidence, or the environment may never have produced it. This section
tests both. Table~\ref{tab:ablation} reports the opportunity-context
ablation; these runs use a different configuration, so absolute
values differ from Table~\ref{tab:recovery}.
\begin{table}[htbp]
\centering
\small
\begin{tabular*}{\columnwidth}{@{\extracolsep{\fill}}lcccc}
\toprule
 & Dodge & Coll. & Aggr. & Skill \\
\midrule
$\rho$ full    & .61 & .39 & .50 & .74 \\
$\rho$ ablated & .61 & .35 & .55 & .59 \\
\bottomrule
\end{tabular*}
\caption{Opportunity-context ablation (Spearman $\rho$, 300 paired
runs).}
\label{tab:ablation}
\end{table}
\paragraph{Opportunity information carries weight.}
Removing what \emph{could have been done} from the records degrades
precisely the two dimensions that depend on opportunities for
expression---Skill falls from .74 to .59 and Collector from .39 to
.35---while Dodge is unchanged.Aggression, by contrast, improves slightly (.50 to .55), suggesting
that the opportunity-related wording was not helping the reader on
this dimension. The
opportunity-aware representation is not a redundant design.

\paragraph{An environment defect, not a reader defect.}
Table~\ref{tab:recovery} raises a question it cannot answer on its
own: why is Collector the weakest dimension for every reader,
handcrafted metrics included? The cause turns out to lie upstream of
reading. In standard levels, reward objects drop from kills, so an
agent that fights little almost never faces the choice of whether to
collect; a strong collector and an indifferent one behave nearly
alike, because neither is ever offered the choice. After constructing
a diagnostic level that actively supplies reward objects, Collector
recoverability rises from .43 to .92 on LLM readings, and
Table~\ref{tab:probe} shows the improvement is not specific to any
reader: every family reads Collector at $\rho \ge .84$ there. To read
out a trait, the environment must first give it an opportunity for
expression, which is also why content that adapts to the player is
not merely a convenience: a fixed level measures only the traits it
happens to afford.

\begin{table}[htbp]
\centering
\small
\setlength{\tabcolsep}{4pt}
\begin{tabular}{llcc}
\toprule
Reader & Type & $\rho$ & MAE \\
\midrule
Handcrafted signature (pursue rate) & metric & \textbf{.92} & --- \\
Rule (calibrated signature) & sup.\ (LOO) & .91 & .097 \\
Linear ridge (17 features) & sup.\ (LOO) & .91 & .121 \\
RBF kernel ridge & sup.\ (LOO) & .84 & .165 \\
BPM prototype (1-NN) & retrieval & .85 & .153 \\
kNN (5-NN) & retrieval & .88 & .165 \\
LLM (GPT-5.6-sol) & LLM & \textbf{.92} & .144 \\
LLM (Qwen3.5-122B-A10B-think) & LLM & .89 & .137 \\
\bottomrule
\end{tabular}
\caption{Collector recovery on the diagnostic probe level
(Spearman $\rho$, $n=30$ per condition). }
\label{tab:probe}
\end{table}

\subsection{RQ4: From Recovery to Adaptation}
Adaptation could improve outcomes for the wrong reason: any level
made easier will kill fewer players, whether or not the profile
driving it is correct. The five arms exist to separate those two
explanations.
The downstream experiment has five arms: adaptation from ground
truth, adaptation from the LLM-inferred profile, uniform difficulty,
a mismatched profile, and plain easing. Each arm is run once per
agent level on a single generated level, and scored on the
flow-matching scale defined above.

\begin{table}[htbp]
\centering
\small
\begin{tabular}{lccccc}
\toprule
Bot & Truth & Inferred & Mismatch & Fixed & Easing \\
\midrule
Novice  & 1.04$^\dagger$ & 1.21$^\dagger$ & 1.17$^\dagger$ & 1.25$^\dagger$ & 1.07$^\dagger$ \\
Low-mid & \textbf{0} & \textbf{0} & 1.15$^\dagger$ & 1.19$^\dagger$ & 1.16$^\dagger$ \\
Mid     & \textbf{0} & 0.33 & 1.06$^\dagger$ & 1.02$^\dagger$ & \textbf{0} \\
High    & \textbf{0} & \textbf{0} & 1.13$^\dagger$ & \textbf{0} & \textbf{0} \\
Expert  & \textbf{0} & 0.19 & 1.19$^\dagger$ & 0.21 & 0.28 \\
\midrule
Survived & 4/5 & 4/5 & 0/5 & 2/5 & 3/5 \\
\bottomrule
\end{tabular}
\caption{Flow-matching score per agent level and arm (lower is
better; \textbf{0} = inside the flow band). $^\dagger$ marks a run
that ended in death, which scores above 1 by construction. One run
per cell.\looseness=-1}
\label{tab:downstream}
\end{table}
With one run per cell, the decimals do not support fine comparison,
but the survival row is unambiguous. The mismatched profile fails at
every level, which is the control that matters: adaptation depends
on the profile being right, not merely on the game being easier.
Ground truth and the inferred profile both survive four of five
runs, and they differ only in degree, the inferred profile landing
outside the flow band at the mid and expert levels where ground
truth lands inside it. That residual gap is the downstream cost of
reading error, and it is smaller than the gap between either of them
and the profile-free arms. Uniform difficulty and plain easing
survive two and three runs, and both fail at the two lowest levels.
The novice level defeats every arm, including ground truth: the
lowest difficulty the generator can assemble is still above what
that agent survives, which is a limit of the generation side rather
than of the reading.

\subsection{Human Study}
Everything so far is synthetic by construction, which is what makes
it checkable and also what limits it. The question the title asks is
whether reading behavior beats asking, when the player is a person.
\begin{figure}[htbp]
  \centering
  \includegraphics[width=\columnwidth]{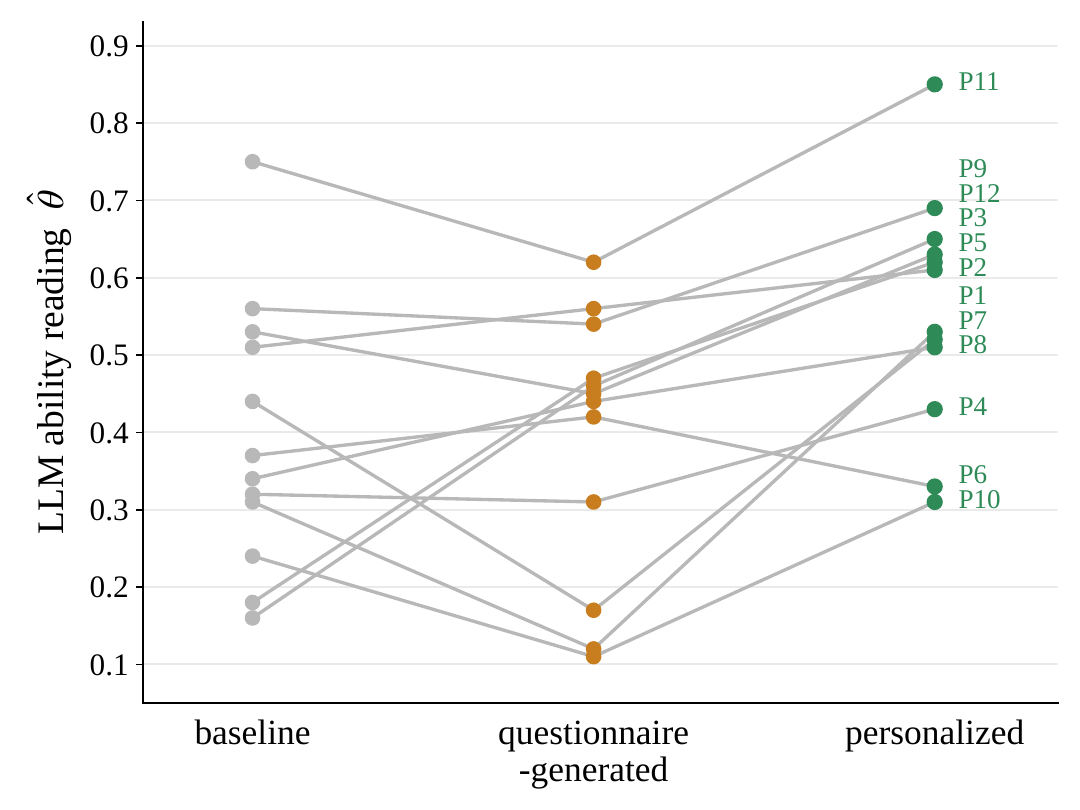}
  \caption{Pilot study ($n=12$). Per-participant LLM ability readings
  under three level-generation conditions; grey lines connect the
  same participant.}
  \label{fig:human-pilot}
\end{figure}
Twelve participants first played a fixed baseline level and then
completed a self-report questionnaire on the experience (appendix).
The questionnaire response and the behavioral record from the same
session were used in parallel to assemble two personalized levels,
which each participant played without being told which was which. Our LLM player model then read the behavioral records from all
sessions. Figure~\ref{fig:human-pilot} plots the resulting ability
readings. 

Readings under the questionnaire-generated condition scatter in both
directions relative to baseline, several participants falling below
it, whereas the behavior-generated condition is higher for every
participant.

\section{Limitations}
This paper builds the apparatus that behavior-based player modeling
has lacked and then uses it. A controlled synthetic population makes
four traits true by construction, and an admission test decides which
of them counts as ground truth at all. An opportunity-aware record
keeps declined choices beside taken ones, and ablating the choice
sets degrades exactly the traits whose expression depends on
opportunity. On this foundation, an LLM player model reads the record
with a fixed example library and no gradient updates, recovering
traits close to a supervised ceiling that presumes labels real
players cannot supply. Closing the loop, the inferred profile
conditions a generator, and in a pilot with twelve participants,
levels built from behavior scored higher than levels built from a
questionnaire.

Three boundaries mark where these results stop. The readers do not
see identical inputs: feature regression consumes continuous
per-session statistics, whereas the reader sees several of them
quantized into bands, so part of the remaining gap belongs to the
transcript rather than to the reading, and a representation that
preserved those quantities is the obvious next thing to test. The
profile is read once from a calibration session and then held fixed,
so nothing here speaks to players whose traits shift as they learn
the game; the replay loop the pipeline provides is the natural place
to track that, and we do not evaluate it. And the synthetic
environment trades the noise of human behavior for control: it is the
trade that makes every claim above checkable, and it is the reason
twelve participants are directional rather than conclusive. Yet every
gap is traced to a cause, and an evaluation that can localize failure
comes closer to the meaning of validation than a demonstration that
can only display success.

\bibliography{beyond_asking}

\begin{thebibliography}{44}
\providecommand{\natexlab}[1]{#1}

\bibitem[{Baker, Saxe, and Tenenbaum(2009)}]{baker2009action}
Baker, C.~L.; Saxe, R.; and Tenenbaum, J.~B. 2009.
\newblock Action Understanding as Inverse Planning.
\newblock \emph{Cognition}, 113: 329--349.

\bibitem[{Brown et~al.(2020)}]{brown2020language}
Brown, T.~B.; et~al. 2020.
\newblock Language Models are Few-Shot Learners.
\newblock In \emph{Advances in Neural Information Processing Systems
  (NeurIPS)}.

\bibitem[{Bunian et~al.(2017)Bunian, Canossa, Colvin, and Seif
  El-Nasr}]{bunian2017modeling}
Bunian, S.; Canossa, A.; Colvin, R.; and Seif El-Nasr, M. 2017.
\newblock Modeling Individual Differences in Game Behavior Using {HMM}.
\newblock In \emph{Proceedings of the AAAI Conference on Artificial
  Intelligence and Interactive Digital Entertainment (AIIDE)}, volume~13,
  158--164.

\bibitem[{Campbell and Fiske(1959)}]{campbell1959convergent}
Campbell, D.~T.; and Fiske, D.~W. 1959.
\newblock Convergent and Discriminant Validation by the Multitrait-Multimethod
  Matrix.
\newblock \emph{Psychological Bulletin}, 56(2): 81--105.

\bibitem[{Chen(2007)}]{chen2007flow}
Chen, J. 2007.
\newblock Flow in Games (and Everything Else).
\newblock \emph{Communications of the ACM}, 50(4): 31--34.

\bibitem[{Cover and Hart(1967)}]{cover1967nearest}
Cover, T.~M.; and Hart, P.~E. 1967.
\newblock Nearest Neighbor Pattern Classification.
\newblock \emph{IEEE Transactions on Information Theory}, 13(1): 21--27.

\bibitem[{Cronbach and Meehl(1955)}]{cronbach1955construct}
Cronbach, L.~J.; and Meehl, P.~E. 1955.
\newblock Construct Validity in Psychological Tests.
\newblock \emph{Psychological Bulletin}, 52(4): 281--302.

\bibitem[{Crowne and Marlowe(1960)}]{crowne1960new}
Crowne, D.~P.; and Marlowe, D. 1960.
\newblock A New Scale of Social Desirability Independent of Psychopathology.
\newblock \emph{Journal of Consulting Psychology}, 24(4): 349--354.

\bibitem[{Csikszentmihalyi(1990)}]{csikszentmihalyi1990flow}
Csikszentmihalyi, M. 1990.
\newblock \emph{Flow: The Psychology of Optimal Experience}.
\newblock Harper and Row.

\bibitem[{Dempster, Laird, and Rubin(1977)}]{dempster1977em}
Dempster, A.~P.; Laird, N.~M.; and Rubin, D.~B. 1977.
\newblock Maximum Likelihood from Incomplete Data via the {EM} Algorithm.
\newblock \emph{Journal of the Royal Statistical Society: Series B
  (Methodological)}, 39(1): 1--22.

\bibitem[{Gallotta et~al.(2024)Gallotta, Todd, Zammit, Earle, Liapis, Togelius,
  and Yannakakis}]{gallotta2024large}
Gallotta, R.; Todd, G.; Zammit, M.; Earle, S.; Liapis, A.; Togelius, J.; and
  Yannakakis, G.~N. 2024.
\newblock Large Language Models and Games: A Survey and Roadmap.
\newblock \emph{IEEE Transactions on Games}.
\newblock ArXiv:2402.18659.

\bibitem[{Goel, Mahmoudi-Nejad, and Guzdial(2024)}]{goel2024labelfree}
Goel, D.; Mahmoudi-Nejad, A.; and Guzdial, M. 2024.
\newblock Label-Free Subjective Player Experience Modelling via Let's Play
  Videos.
\newblock In \emph{Proceedings of the AAAI Conference on Artificial
  Intelligence and Interactive Digital Entertainment}, volume~20, 46--53.

\bibitem[{Hoerl and Kennard(1970)}]{hoerl1970ridge}
Hoerl, A.~E.; and Kennard, R.~W. 1970.
\newblock Ridge Regression: Biased Estimation for Nonorthogonal Problems.
\newblock \emph{Technometrics}, 12(1): 55--67.

\bibitem[{Holmg{\aa}rd et~al.(2019)Holmg{\aa}rd, Green, Liapis, and
  Togelius}]{holmgard2019automated}
Holmg{\aa}rd, C.; Green, M.~C.; Liapis, A.; and Togelius, J. 2019.
\newblock Automated Playtesting with Procedural Personas through {MCTS} with
  Evolved Heuristics.
\newblock \emph{IEEE Transactions on Games}, 11(4): 352--362.

\bibitem[{Holmg{\aa}rd et~al.(2014)Holmg{\aa}rd, Liapis, Togelius, and
  Yannakakis}]{holmgard2014evolving}
Holmg{\aa}rd, C.; Liapis, A.; Togelius, J.; and Yannakakis, G.~N. 2014.
\newblock Evolving Personas for Player Decision Modeling.
\newblock In \emph{Proceedings of the 2014 IEEE Conference on Computational
  Intelligence and Games (CIG)}, 1--8.

\bibitem[{Holmg{\aa}rd et~al.(2015)Holmg{\aa}rd, Liapis, Togelius, and
  Yannakakis}]{holmgard2015monte}
Holmg{\aa}rd, C.; Liapis, A.; Togelius, J.; and Yannakakis, G.~N. 2015.
\newblock Monte-Carlo Tree Search for Persona Based Player Modeling.
\newblock In \emph{Proceedings of the AAAI Conference on Artificial
  Intelligence and Interactive Digital Entertainment (AIIDE)}, volume~11,
  8--14.

\bibitem[{Hunicke(2005)}]{hunicke2005dda}
Hunicke, R. 2005.
\newblock The Case for Dynamic Difficulty Adjustment in Games.
\newblock In \emph{Proceedings of the 2005 ACM SIGCHI International Conference
  on Advances in Computer Entertainment Technology (ACE)}, 429--433.

\bibitem[{John and Srivastava(1999)}]{john1999big}
John, O.~P.; and Srivastava, S. 1999.
\newblock The {Big-Five} Trait Taxonomy: History, Measurement, and Theoretical
  Perspectives.
\newblock In \emph{Handbook of Personality: Theory and Research}, 102--138.
  Guilford Press, 2nd edition.

\bibitem[{Liu et~al.(2022)Liu, Shen, Zhang, Dolan, Carin, and
  Chen}]{liu2022makes}
Liu, J.; Shen, D.; Zhang, Y.; Dolan, B.; Carin, L.; and Chen, W. 2022.
\newblock What Makes Good In-Context Examples for {GPT-3}?
\newblock In \emph{Proceedings of Deep Learning Inside Out (DeeLIO), ACL
  Workshop}.

\bibitem[{Lopes, Fachada, and Fonseca(2025)}]{lopes2025closing}
Lopes, P.; Fachada, N.; and Fonseca, M. 2025.
\newblock Closing the Loop: A Systematic Review of Experience-Driven Game
  Adaptation.
\newblock ArXiv:2505.01351.

\bibitem[{Maharjan et~al.(2025)Maharjan, Jin, Zhu, and
  Kenne}]{maharjan2025psychometric}
Maharjan, J.; Jin, R.; Zhu, J.; and Kenne, D. 2025.
\newblock Psychometric Evaluation of Large Language Model Embeddings for
  Personality Trait Prediction.
\newblock \emph{Journal of Medical Internet Research}, 27: e75347.

\bibitem[{Mislevy, Steinberg, and Almond(2003)}]{mislevy2003ecd}
Mislevy, R.~J.; Steinberg, L.~S.; and Almond, R.~G. 2003.
\newblock On the Structure of Educational Assessments.
\newblock \emph{Measurement: Interdisciplinary Research and Perspectives},
  1(1): 3--62.

\bibitem[{Mortazavi, Moradi, and Vahabie(2024)}]{mortazavi2024dda}
Mortazavi, F.; Moradi, H.; and Vahabie, A.-H. 2024.
\newblock Dynamic Difficulty Adjustment Approaches in Video Games: A Systematic
  Literature Review.
\newblock \emph{Multimedia Tools and Applications}, 83.

\bibitem[{Ng and Russell(2000)}]{ng2000algorithms}
Ng, A.~Y.; and Russell, S. 2000.
\newblock Algorithms for Inverse Reinforcement Learning.
\newblock In \emph{Proceedings of the Seventeenth International Conference on
  Machine Learning (ICML)}, 663--670.

\bibitem[{Nisbett and Wilson(1977)}]{nisbett1977telling}
Nisbett, R.~E.; and Wilson, T.~D. 1977.
\newblock Telling More Than We Can Know: Verbal Reports on Mental Processes.
\newblock \emph{Psychological Review}, 84(3): 231--259.

\bibitem[{{OpenAI}(2026)}]{openai2026gpt56}
{OpenAI}. 2026.
\newblock {GPT-5.6}: Frontier Intelligence That Scales with Your Ambition.

\bibitem[{Peters, Cerf, and Matz(2024)}]{peters2024large}
Peters, H.; Cerf, M.; and Matz, S.~C. 2024.
\newblock Large Language Models Can Infer Personality from Free-Form User
  Interactions.
\newblock arXiv:2405.13052.

\bibitem[{Podsakoff et~al.(2003)Podsakoff, MacKenzie, Lee, and
  Podsakoff}]{podsakoff2003common}
Podsakoff, P.~M.; MacKenzie, S.~B.; Lee, J.-Y.; and Podsakoff, N.~P. 2003.
\newblock Common Method Biases in Behavioral Research: A Critical Review of the
  Literature and Recommended Remedies.
\newblock \emph{Journal of Applied Psychology}, 88(5): 879--903.

\bibitem[{{Qwen Team}(2026)}]{qwen2026qwen35}
{Qwen Team}. 2026.
\newblock {Qwen3.5-122B-A10B}.

\bibitem[{Sch{\"o}lkopf and Smola(2002)}]{scholkopf2002learning}
Sch{\"o}lkopf, B.; and Smola, A.~J. 2002.
\newblock \emph{Learning with Kernels: Support Vector Machines, Regularization,
  Optimization, and Beyond}.
\newblock Cambridge, MA: MIT Press.

\bibitem[{Shaker, Togelius, and Nelson(2016)}]{shaker2016procedural}
Shaker, N.; Togelius, J.; and Nelson, M.~J. 2016.
\newblock \emph{Procedural Content Generation in Games}.
\newblock Springer.

\bibitem[{Shergadwala, Teng, and Seif El-Nasr(2021)}]{shergadwala2021can}
Shergadwala, M.~N.; Teng, Z.; and Seif El-Nasr, M. 2021.
\newblock Can We Infer Player Behavior Tendencies from a Player's
  Decision-Making Data? {Integrating} Theory of Mind to Player Modeling.
\newblock In \emph{Proceedings of the AAAI Conference on Artificial
  Intelligence and Interactive Digital Entertainment (AIIDE)}, volume~17,
  195--202.

\bibitem[{Shute and Ventura(2013)}]{shute2013stealth}
Shute, V.; and Ventura, M. 2013.
\newblock \emph{Stealth Assessment: Measuring and Supporting Learning in Video
  Games}.
\newblock MIT Press.

\bibitem[{Sudhakaran et~al.(2023)Sudhakaran, Gonz{\'a}lez-Duque, Freiberger,
  Glanois, Najarro, and Risi}]{sudhakaran2023mariogpt}
Sudhakaran, S.; Gonz{\'a}lez-Duque, M.; Freiberger, M.; Glanois, C.; Najarro,
  E.; and Risi, S. 2023.
\newblock MarioGPT: Open-Ended Text2Level Generation through Large Language
  Models.
\newblock In \emph{Advances in Neural Information Processing Systems
  (NeurIPS)}.
\newblock ArXiv:2302.05981.

\bibitem[{Talts et~al.(2020)Talts, Betancourt, Simpson, Vehtari, and
  Gelman}]{talts2020validating}
Talts, S.; Betancourt, M.; Simpson, D.; Vehtari, A.; and Gelman, A. 2020.
\newblock Validating Bayesian Inference Algorithms with Simulation-Based
  Calibration.
\newblock arXiv:1804.06788.

\bibitem[{Tett and Burnett(2003)}]{tett2003personality}
Tett, R.~P.; and Burnett, D.~D. 2003.
\newblock A Personality Trait-Based Interactionist Model of Job Performance.
\newblock \emph{Journal of Applied Psychology}, 88(3): 500--517.

\bibitem[{Todd et~al.(2023)Todd, Earle, Nasir, Green, and
  Togelius}]{todd2023llm}
Todd, G.; Earle, S.; Nasir, M.~U.; Green, M.~C.; and Togelius, J. 2023.
\newblock Level Generation Through Large Language Models.
\newblock In \emph{Proceedings of the 18th International Conference on the
  Foundations of Digital Games (FDG)}.

\bibitem[{Woods et~al.(2015)Woods, Wyma, Yund, Herron, and
  Reed}]{woods2015factors}
Woods, D.~L.; Wyma, J.~M.; Yund, E.~W.; Herron, T.~J.; and Reed, B. 2015.
\newblock Factors Influencing the Latency of Simple Reaction Time.
\newblock \emph{Frontiers in Human Neuroscience}, 9: 131.

\bibitem[{Yannakakis et~al.(2013)Yannakakis, Spronck, Loiacono, and
  Andr{\'e}}]{yannakakis2013player}
Yannakakis, G.~N.; Spronck, P.; Loiacono, D.; and Andr{\'e}, E. 2013.
\newblock Player Modeling.
\newblock In \emph{Artificial and Computational Intelligence in Games},
  volume~6 of \emph{Dagstuhl Follow-Ups}, 45--59.

\bibitem[{Yannakakis and Togelius(2011)}]{yannakakis2011experience}
Yannakakis, G.~N.; and Togelius, J. 2011.
\newblock Experience-Driven Procedural Content Generation.
\newblock \emph{IEEE Transactions on Affective Computing}, 2(3): 147--161.

\bibitem[{Yannakakis and Togelius(2018)}]{yannakakis2018artificial}
Yannakakis, G.~N.; and Togelius, J. 2018.
\newblock \emph{Artificial Intelligence and Games}.
\newblock Springer.

\bibitem[{Ye et~al.(2026)Ye, Jin, Xie, Zhang, and Song}]{ye2026large}
Ye, H.; Jin, J.; Xie, Y.; Zhang, X.; and Song, G. 2026.
\newblock Large Language Model Psychometrics: A Systematic Review of
  Evaluation, Validation, and Enhancement.
\newblock arXiv:2505.08245.

\bibitem[{Zhu, Jin, and Coifman(2025)}]{zhu2025can}
Zhu, J.; Jin, R.; and Coifman, K.~G. 2025.
\newblock Can LLMs Infer Personality from Real World Conversations?
\newblock arXiv:2507.14355.

\bibitem[{Zhu et~al.(2025)Zhu, Maharjan, Li, Coifman, and
  Jin}]{zhu2025evaluating}
Zhu, J.; Maharjan, J.; Li, X.; Coifman, K.~G.; and Jin, R. 2025.
\newblock Evaluating LLM Alignment on Personality Inference from Real-World
  Interview Data.
\newblock arXiv:2509.13244.

\end{thebibliography}

\clearpage
\appendix
\renewcommand{\thefigure}{A\arabic{figure}}
\renewcommand{\thetable}{A\arabic{table}}
\renewcommand{\theequation}{A\arabic{equation}}
\renewcommand{\thealgorithm}{A\arabic{algorithm}}
\setcounter{figure}{0}
\setcounter{table}{0}
\setcounter{equation}{0}
\setcounter{secnumdepth}{2}
\renewcommand{\thesection}{\Alph{section}}
\renewcommand{\thesubsection}{\Alph{section}.\arabic{subsection}}
\setcounter{section}{0}

\section{Synthetic-Player Policy}
\label{app:policy}

The synthetic player (Bot) is a real-time decision maker running at
sixty frames per second, decoupled from rendering and driven by a
single random seed: given the four trait parameters
$\theta\in[0,1]^4$ and the seed, the trajectory is reproducible frame
by frame. Ground truth is therefore the generative parameter, not a
post-hoc annotation. The policy has three layers: objective
arbitration, a movement layer, and an action layer.

\subsection{Trait Parameterization (Constants)}
\label{app:policy-knobs}
Each $\theta_k$ enters exactly one channel: Dodge scales the
movement layer, Collector and Aggression enter the
arbitration intent gate, and Skill sets the action layer's
per-opportunity use probability
$P(\text{use}\mid\text{opportunity})=\theta_{\text{skill}}$.

Both preference dimensions enter through the intent gate,
\begin{equation}
\text{pursue}_k \iff u_t<\theta_k,\quad u_t\sim U(0,1)\ \text{resampled every }2\,\text{s},
\label{eq:gate-app}
\end{equation}
so the expected share of pursuit time equals $\theta_k$. The interval
is re-rolled every $2$\,s rather than held fixed for a whole pursuit,
which keeps a high-aggression bot from locking onto one tanky enemy and
washing out the signal. Both also shrink the safety margin required
before pursuit,
\begin{equation}
T_{\text{safe}}(\theta_k)=T_0\,(1-0.5\,\theta_k),
\label{eq:risk-app}
\end{equation}
and the reaction interval is the dominant Dodge channel,
$\Delta t(\theta_{\text{dodge}})=420-340\,\theta_{\text{dodge}}$\,ms.

\subsection{Movement Layer: Danger Field and Time-to-Collision}
\label{app:geometry}
At each decision step the agent samples candidate destinations and
scores each with a \emph{danger field}, moving toward the reachable
point of lowest score subject to a time-to-collision (TTC) veto. Both
quantities are computed analytically, so a fast hazard cannot slip
between time samples.

\paragraph{Danger field.}
For candidate point $p$ and hazard $b$ with position--velocity
$(\mathbf{d},\mathbf{v})$, $\mathbf{d}=(b_x-p_x,b_y-p_y)$, the time of
closest approach over the foresight horizon $H_d$ is
\begin{equation}
t^{*}=\mathrm{clip}\!\Big(-\tfrac{\mathbf{d}\cdot\mathbf{v}}{\lVert\mathbf{v}\rVert^2},\,0,\,H_d\Big),
\label{eq:tstar-app}
\end{equation}
the closest-approach clearance is $m_b=\lVert\mathbf{d}+t^{*}\mathbf{v}\rVert-r_b$,
and a hazard contributes only if $m_b<R_d$ (danger radius $R_d=110$\,px):
\begin{equation}
c_b=\frac{R_d-m_b}{R_d}\quad(\text{unclamped for }m_b<0).
\label{eq:cb-app}
\end{equation}
Leaving $c_b$ unclamped gives a continuous outward gradient out of a
hazard body. The field is \emph{max-dominated}, not summed:
\begin{equation}
\mathrm{Danger}(p)=c_{\max}+0.15\big(\textstyle\sum_b c_b-c_{\max}\big),\quad c_{\max}=\max_b c_b,
\label{eq:danger-app}
\end{equation}
so the acute-threat gradient survives at any bullet density; a pure sum
saturates under dense fire and the agent freezes.

\paragraph{Time to collision.}
For the hero at relative position $\mathbf{r}_p$ and velocity
$\mathbf{r}_v$ against combined radius $r$, TTC is the smaller
nonnegative root of $\lVert\mathbf{r}_p+t\,\mathbf{r}_v\rVert=r$:
\begin{equation}
\mathrm{TTC}=\frac{-b-\sqrt{b^2-4ac}}{2a},\ \ a=\lVert\mathbf{r}_v\rVert^2,\ b=2\,\mathbf{r}_p\!\cdot\!\mathbf{r}_v,\ c=\lVert\mathbf{r}_p\rVert^2-r^2,
\label{eq:ttc-app}
\end{equation}
returning $+\infty$ when the hazard moves apart or does not intersect.
A candidate direction whose soonest bullet-or-boundary collision falls
below $\mathrm{TTC}_{\min}=0.18$\,s is vetoed; if all directions are
unsafe, the agent takes the one with the largest TTC. Hero and enemy
collision radii are $0.175\,w_{\text{hero}}$ and
$0.4\,w_{\text{enemy}}+0.175\,w_{\text{hero}}$ respectively
($w_{\text{hero}}\approx68$\,px).

\subsection{Decision Loop (Pseudocode)}
\label{app:policy-alg}
\begin{algorithm}[t]
\caption{Synthetic-player decision loop (one frame)}
\label{alg:bot}
\begin{algorithmic}[1]
\REQUIRE traits $\theta$, state $s_t$, seed-driven $\mathrm{rng}$
\IF{$t-t_{\text{roll}}\ge 2000$\,ms}
  \STATE $u_{\text{coll}},u_{\text{aggr}}\leftarrow\mathrm{rng}()$;\ \ $t_{\text{roll}}\leftarrow t$
\ENDIF
\STATE $\text{obj}\leftarrow$ earliest-deadline objective with $u<\theta$ (Eq.~\ref{eq:gate-app}), gated by $T_{\text{safe}}$ (Eq.~\ref{eq:risk-app})
\STATE sample candidate points; score each by $\mathrm{Danger}(\cdot)$ (Eq.~\ref{eq:danger-app})
\STATE $\text{dir}\leftarrow\arg\min$ danger among directions with $\mathrm{TTC}\ge0.18$\,s (Eq.~\ref{eq:ttc-app}); else max-TTC
\IF{ability opportunity present \AND $\mathrm{rng}()<\theta_{\text{skill}}$}
  \STATE fire missile / activate berserk
\ENDIF
\STATE move toward $\text{obj}$ if chosen, else toward $\text{dir}$
\end{algorithmic}
\end{algorithm}

\subsection{Alternative Policy Family}
A second \emph{utility family} replaces the heuristics above with
utility scoring over candidate actions; the same parameters produce
similar tendencies. A full cross-family recovery evaluation is left to
future work.

\section{Ground-Truth Admission}
\label{app:admission}

\subsection{Signature Metrics}
Each dimension is validated against a held-out \emph{signature metric}
that plays no role in defining the parameter:
Dodge~$\to$ wave-phase hits per exposed minute (expect $\downarrow$);
Collector~$\to$ XP pursue rate $=$ picked/spawned (expect $\uparrow$);
Aggression~$\to$ mean flight-$y$ position (expect $\downarrow$;
lower $y$ means the ship pressed forward toward the enemies---the
strongest single feature effect in the benchmark, $\rho\approx-.91$);
Skill~$\to$ ability attempts per exposed (non-berserk) minute
(expect $\uparrow$).

\subsection{Criteria and Thresholds}
Each parameter is swept over $L=6$ levels
$\theta\in\{0,.2,.4,.6,.8,1.0\}$ with $5$ seeds per level
($n=30$ per dimension). All metrics, directions, and thresholds were
frozen before any model response was observed. A parameter is admitted
iff:
\begin{itemize}
\item[C1] \textbf{Monotonicity (convergent validity).} Spearman $\rho$
  between level and its signature metric satisfies $\rho\ge 0.6$ with
  the expected sign and no significant non-monotone segment.
\item[C2] \textbf{Isolation (no systematic drift).} The full
  cross-dimension matrix of Spearman correlations between the swept
  level and \emph{every} axis's signature metric is computed; the
  diagonal must dominate, and any off-diagonal $|\rho|>0.5$ is flagged
  as cross-talk.
\item[C3] \textbf{Separability.} Adjacent levels, plus the
  $(0.4,1.0)$ and $(0.6,1.0)$ pairs, are compared by an exact
  Mann--Whitney permutation test ($5$ vs.\ $5$, all $252$ splits) with
  a directed $\mathrm{AUC}=U/(n_1 n_2)$; a pair with
  $\mathrm{AUC}<0.7$ is flagged indistinguishable and greedily merged
  into effective levels. A one-way ANOVA $F$ and $\eta^2$ across level
  groups are reported as signal-to-noise summaries but are \emph{not}
  used as an admission gate.
\end{itemize}
Configurations that fail are reported as rejected or non-identifiable.

\section{Behavioral Records}
\label{app:records}

\subsection{Format}
Each session serializes to plain text containing: decision-moment
lines (state $+$ the action taken, with declined opportunities on file
alongside taken ones), combat-event lines, ten-second windowed
summaries (spawn/kill/escape counts, orbs spawned/picked/expired,
grazes, positional band), and a closing whole-session tally. The
summaries supply the \emph{denominators} against which willingness is
judged (e.g., orbs \emph{spawned} vs.\ \emph{picked}).

\subsection{Leakage Control}
\label{app:leak}
No trait name, internal metric key, provenance field, or precomputed
statistic ever appears in a record. This is enforced by a denylist
check, \texttt{assertNoLeakage}, run after every serialization; a hit
raises a hard error. The denylist covers (i) trait/axis names
(\texttt{dodge\_skill}, \texttt{collector}, \texttt{aggression},
\texttt{skill\_usage}, \dots); (ii) metric keys from the measurement
pipeline (\texttt{hits\_rate}, \texttt{pursue\_rate},
\texttt{kill\_conversion}, \dots); (iii) provenance/labels
(\texttt{oracle}, \texttt{ground truth}, \texttt{generated\_spec},
\texttt{calibration}); (iv) statistics machinery (\texttt{z-score},
\texttt{sigmoid}, \texttt{percentile}, \texttt{baseline}); and (v) the
bot's internal decision flags (\texttt{wantKill}, \texttt{wantOrb},
\texttt{safetyGateOpen}). Because a bot session and a human session are
serialized by the same procedure into the same leakage-free schema,
the reader operates on the same kind of object regardless of origin;
the parameter $\theta$ generates behavior upstream and never enters
what is read.

\subsection{Excerpt}
The lines below are verbatim from a real \emph{human} session
(alpha level, $123.7$\,s, outcome WIN):
\begin{quote}\footnotesize\ttfamily
[t=23.6s] DECISION engage: hp=175 bombs=0\\
\hspace*{1em}nearestBullet=116px danger=0 enemies=3\\
\hspace*{1em}berserkActive=no -> EVADED\\
{}[t=25.2s] BERSERK activated (hp=175,\\
\hspace*{1em}bullets onscreen=16, nearest=60px, enemies=6)\\
{}[t=29.7s] DECISION xp-orb: hp=175 bombs=0\\
\hspace*{1em}nearestBullet=143px danger=0 enemies=5\\
\hspace*{1em}berserkActive=yes -> AVOIDED\\
{}[t=30.0s] SUMMARY 20-30s: enemies spawned=11 killed=5\\
\hspace*{1em}escaped=4 | xp orbs spawned=5 picked=2 | grazes=7\\
{}[t=35.7s] DECISION xp-orb: hp=175 bombs=0\\
\hspace*{1em}nearestBullet=173px danger=0 enemies=8\\
\hspace*{1em}berserkActive=no -> AVOIDED
\end{quote}
The two \texttt{xp-orb $\to$ AVOIDED} lines are collection
opportunities that were present but declined; on file alongside the
\texttt{spawned=5 picked=2} denominator, ``did not'' can be told apart
from ``could not''.

\section{The Reader}
\label{app:reader}

\subsection{Prompt}
The reader is a single fixed function from a transcript to a
four-dimensional estimate. It is a prompt, not a trained model: the same
prompt is used for every session and every backbone, and it is frozen
before any result is read. The full text is reproduced verbatim below.
It has three parts: a description of the game and the reading task, a
set of scoring rules that fix how observations map to numbers, and the
required response format. The transcript is appended after it as the
user message.

\begin{quote}\footnotesize
\ttfamily
You are an expert analyst of video-game player behavior. You will
receive a behavioral transcript of one run of a vertical-scrolling
shoot-em-up ("shmup") played by an unknown player, and you must
profile that player.\\[2pt]
THE GAME: The player pilots a ship over a vertically scrolling
battlefield, auto-firing upward at enemies that enter from the top.
Enemies shoot bullets; getting hit costs HP and the run ends in
defeat at 0 HP. Destroyed enemies drop XP orbs that disappear if not
collected; flying close to an orb collects it. Gift drops give weapon
upgrades (blue) or missile refills (red). The player has two limited
abilities: MISSILES (a screen-clearing burst, limited stock) and
BERSERK (a short invulnerable power window on a cooldown, triggered
at will). A "graze" is a bullet passing very close without hitting.
The run ends by defeating the boss (WIN), dying (LOSS), or hitting a
session time cap (TIMEOUT).\\[2pt]
THE TRANSCRIPT contains: a header (level, duration, outcome, final
HP, missiles left); DECISION lines capturing the full game state at
notable choice points and the action the player took; combat beats
(hits taken, missile launches, berserk activations, pickups, boss
phases); and per-10-second SUMMARY lines (enemies
spawned/killed/escaped, xp orbs spawned/picked/expired, grazes) plus
POSITION lines (vertical band the ship flew in --- FRONT is up near
the enemies, BACK is the safe bottom --- and how close enemy bullets
got).\\[2pt]
YOUR TASK: estimate four independent traits of this player. Each is a
number in [0,1] where 0 means the extreme low end of the player
population and 1 the extreme high end. Judge each trait purely from
observable behavior in the transcript:\\[2pt]
- "dodge\_skill" --- how skillfully the player avoids getting hit
while under fire. High: stays unhit even when many bullets are on
screen and passing close; keeps control in dense patterns; survives
long stretches without losing HP. Low: repeatedly takes hits whenever
bullets are around, bleeds HP quickly under pressure. Judge relative
to the danger actually faced.\\[2pt]
- "collector" --- the player's drive to gather rewards. High: chases
down XP orbs and gift drops before they expire, detours toward
pickups even from a comfortable position, ends up collecting most of
what dropped. Low: ignores orbs and gifts, lets most rewards drift
away uncollected.\\[2pt]
- "aggression" --- the player's drive to seek out and destroy
enemies. High: pushes toward enemies, holds position to finish kills,
keeps fighting rather than backing off, lets few enemies escape. Low:
hangs back passively, avoids confrontation, lets many enemies leave
alive.\\[2pt]
- "skill\_usage" --- the player's willingness to actually spend the
limited abilities (missiles and berserk) when threats or
opportunities arise, rather than hoarding them. High: uses berserk
and missiles readily and repeatedly. Low: finishes the run with
abilities unspent even in dangerous moments.\\[2pt]
These traits are independent. Base everything ONLY on the transcript;
do not assume a trait value because of the outcome alone.\\[4pt]
SCORING RULES --- your numbers must follow these frequency
definitions, not a general impression:\\[2pt]
- "skill\_usage": the FRACTION of clear opportunities the player
actually spent an ability on --- never the raw count of uses. Two
denominators are visible: (1) DECISION bomb lines record each missile
offer and whether it was USED or HELD --- the USED share is the most
direct read; (2) berserk recharges $\sim$10s start-to-start, so
$\sim$6 activations fit in a minute of sustained combat:
$\sim$5-6/min = 0.9-1.0, $\sim$3/min = 0.4-0.5, 1-2/min while enemies
stay onscreen = 0.0-0.3. If the screen was mostly empty, judge from
the missile DECISION share alone.\\[2pt]
- "aggression": judge from NORMAL-flight positioning (POSITION lines)
and engagement: parked deep in BACK band letting most enemies leave =
0.0-0.2; MID band, engaging targets that come close = $\sim$0.5;
pressing toward FRONT and hunting most enemies down = 0.8-1.0. Kills
made while invincible say little about aggression.\\[2pt]
- "dodge\_skill": judge from the RATE of HIT lines RELATIVE TO bullet
exposure, never from the final HP or whether the run was a LOSS.
Taking few hits across many windows of close, dense fire = 0.8-1.0;
getting hit in most exposed windows = 0.0-0.2; roughly one hit per
couple of minutes of real pressure = $\sim$0.5. Do NOT lower this
just because HP ran low or the run ended in defeat.\\[2pt]
Judge "collector" from its definition above as usual.\\[4pt]
Respond with a single JSON object, nothing else:\\
\{"dodge\_skill": n, "collector": n, "aggression": n,
"skill\_usage": n, "rationale": "2-4 sentences citing concrete
transcript evidence"\}
\end{quote}

\noindent\textbf{Two clauses added for the human study.} The human
study adds two clauses to the skill-usage rule. First, a
\texttt{DECISION bomb} line with \texttt{bombs=0} is not a declined
offer---there was no missile to spend---so it is left out of the
USED-share denominator; running the stock down to zero by spending
counts \emph{for} high skill usage, not against it. Second, the berserk
rate is computed directly (activations $\div$ minutes) and the scale
applied from that number, rather than judged by impression.

\subsection{Few-Shot Retrieval}
Two kinds of labeled evidence are placed in context: a table of
$N{=}100$ rows mapping behavioral metrics to verified parameters, and
the $k{=}2$ most similar sessions retrieved by \emph{z}-scored
Euclidean distance over behavioral features, each paired with its
verified profile. Two leakage guards apply: the query run itself
(matched by run identifier) is never retrieved, and any exact feature
duplicate (distance $\approx0$) is dropped. All labeled evidence
originates from a fixed calibration level, so labeling is paid once.

\subsection{Decoding and Repeats}
The reader uses the provider defaults (no explicit temperature or
top-$p$) with a JSON-object response format; each session is read
$3$ times independently and the per-dimension scores averaged, with the
session as the unit of statistical analysis. The reported backbones are
\texttt{gpt-5.6-sol} and \texttt{Qwen3.5-122B-A10B-think}.

\section{Baselines}
\label{app:baselines}

We compare the reader against five baselines, all evaluated on the same
frozen $90$-session test set. The supervision available to the two sides
is deliberately \emph{not} equal. Every baseline is fitted on the $210$
labeled training sessions, drawn from the same distribution as the test
set; the reader performs no gradient updates and consumes only a fixed,
offline example library placed in context. The supervised baselines are
therefore best read as an instrument rather than as competitors: they
measure how much trait information the records contain when
in-distribution labels are available, a condition that cannot hold for
real players whose true traits are unobservable. Four of the five read
the same $17$ $z$-scored behavioral features
(Table~\ref{tab:features}); one (EmbReg) reads the transcript text
itself, exactly as the reader does.

\paragraph{Rule --- handcrafted signatures.}
The simplest baseline, and the ``obvious feature, read directly''
reference. Each trait is predicted from one hand-picked statistic---the
same kind of held-out signature used to admit that trait in
Section~\ref{app:admission} (e.g.\ hits per exposed minute for Dodge,
XP pursue rate for Collector, mean flight-$y$ for Aggression, ability
attempts per minute for Skill)---rescaled monotonically to $[0,1]$. It
has no learned parameters.

\paragraph{FeatReg --- feature kernel ridge.}
The strongest baseline overall. It fits one kernel ridge regressor
per trait, with an RBF kernel over the $17$ features; the ridge penalty
$\lambda$ is chosen per trait by inner $4$-fold cross-validation over
$\lambda\in\{1,10,100,1000\}$. A plain linear-ridge version is only
slightly weaker (macro $\rho$ $.75$ vs.\ $.77$), so most of the accuracy
comes from the features rather than the kernel. FeatReg is the ceiling
referred to above: given same-distribution labels, it measures how much
trait signal the records actually contain.

\paragraph{EmbReg --- embedding kernel ridge.}
The same kernel-ridge setup, but over the transcript \emph{text} instead
of the engineered features. Each transcript is embedded with OpenAI
\texttt{text-embedding-3-small} ($1536$-d) and mean-pooled to one
vector, then regressed to the four traits with the same $\lambda$
selection. Because it reads exactly what the reader reads, it isolates
whether an off-the-shelf text embedding already carries the trait signal
the reader extracts.

\paragraph{BPM 1-NN --- behavioral-prototype matching.}
A nonparametric memory baseline: each test session is matched to its
single nearest training session in the $z$-scored $17$-feature space,
and takes that neighbor's four labels as its prediction. It asks whether
trait recovery is just ``retrieve the most similar labeled player.''

\paragraph{GMM --- soft and hard clustering.}
Fits eight Gaussian clusters with diagonal covariance to the training
features by EM (seed $7$, $200$ iterations, variance floor $10^{-2}$);
each cluster stores the mean trait vector of its member sessions. The
\emph{soft} variant predicts a responsibility-weighted average of those
cluster trait-means; the \emph{hard} variant uses only the single most
likely cluster. It tests whether a handful of behavioral ``player
types'' is enough to place a session.

\begin{table}[t]
\centering \footnotesize
\begin{tabular}{@{}p{0.94\columnwidth}@{}}
\toprule
\textbf{FeatReg feature set ($17$)} \\
\midrule
hits\_rate, graze\_rate, mean\_bullet\_dist, gun\_kill\_conversion,
kill\_conversion, kills\_per\_min, escape\_share, xp\_pursue\_rate,
xp\_expired\_share, gift\_rate, missile\_rate, berserk\_rate,
ability\_rate, mean\_y, std\_y, front\_share, invincible\_share \\
\bottomrule
\end{tabular}
\caption{Behavioral features used by the supervised feature baselines.}
\label{tab:features}
\end{table}

\section{Difficulty Model}
\label{app:difficulty}

\subsection{Content Difficulty Coefficient $D_L$}
$D_L=\sigma\!\big(\sum_i w_i\ln(x_i/x_i^{\mathrm{ref}})\big)$ measures
how much pressure a level's content puts on the player. Each content
feature $x_i$ is compared to its value on a standard calibration level
as a log ratio $\ln(x_i/x_i^{\mathrm{ref}})$; the weighted sum of those
ratios is passed through a sigmoid, anchored so the standard level sits
at $D_L=0.5$. The weights $w_i$ are fixed in advance from measured
feature elasticities, not learned, so $D_L$ is an interpretable
content-pressure coefficient rather than a fitted success predictor
(Table~\ref{tab:dlweights}).
\begin{table}[t]
\centering \footnotesize
\begin{tabular}{@{}lr@{\hskip 2em}lr@{}}
\toprule
Feature & $w_i$ & Feature & $w_i$ \\
\midrule
trackerShare      & $.25$ & effectiveHp     & $.10$ \\
bulletsPerSec     & $.20$ & dangerExposure  & $.10$ \\
spawnRatePerMin   & $.15$ & bulletSpeedMul  & $.00$ \\
peakConcurrent    & $.15$ & reliefPerMin    & $-.05$ \\
\bottomrule
\end{tabular}
\caption{$D_L$ prior weights ($\Sigma|w|=1$). \texttt{bulletSpeedMul}
is measured non-monotone and excluded; \texttt{reliefPerMin} is a
relief term (negative).}
\label{tab:dlweights}
\end{table}

\subsection{Target Surface $g(\cdot)$}
The target difficulty $D^{*}=g(\hat\theta)$ is set from the estimated
Dodge and Aggression. The mapping runs through a response surface
$\text{hits/min}=f(\text{dodge},\text{fireMul})$ fitted on $167$ runs
(a dual-difficulty battery); the fire multiplier is clamped to
$[0.18,3.0]$.

\subsection{Flow-Matching Outcome Scale}
\label{app:flow}
Downstream outcomes are scored on a flow-matching scale (lower is
better). The score is read from a behavioral quantity the generator does
not control---hits per minute, $h$. Let $[L,H]$ be a flow band frozen in
advance from mid-level agents, $C$ the session-length cap, and $T$ the
run duration:
\begin{equation}
\text{score}=
\begin{cases}
1+\dfrac{C-T}{C} & \text{death } (T<C,\ \text{no win})\\[4pt]
\min\!\big(1,\tfrac{h-H}{H}\big) & \text{survived},\ h>H\ \text{(overload)}\\[4pt]
\min\!\big(1,\tfrac{L-h}{L}\big) & \text{survived},\ h<L\ \text{(boredom)}\\[2pt]
0 & \text{survived},\ L\le h\le H\ \text{(flow)}.
\end{cases}
\label{eq:flow-app}
\end{equation}
The jump at $1$ is deliberate: hits per minute cannot tell a fast
death from a short clean run, so a death is placed above any survived
outcome and ranked by how early it ended. The band used for the
downstream flow-matching score is $[L,H]=[2.8,7.3]$ hits/min, frozen
in advance from mid-level reference agents (mean $5.07\pm1.5\sigma$);
it is distinct from the generation-side targeter band and is never
used to control generation.

\section{Generation Pipeline}
\label{app:generation}

\subsection{Director and Translation}
\label{app:gen-director}
A recovered profile becomes a playable level in four stages. The first
two are done by a language model (which we call the \emph{director});
the last two are deterministic code.

\emph{(1) Set the target.} The profile is turned into a director input.
Alongside the profile, this input states the target difficulty as an
average scene intensity---a single number $[\text{Intensity }X/10]$
derived from $D^{*}$. Stating the difficulty this way makes the director
generate \emph{toward} a difficulty, instead of free-styling from the
profile alone.

\emph{(2) Draft the level.} The director (\texttt{gpt-5.6-sol}) writes a
structured draft: a world, an enemy roster, a boss, and $5$--$9$
\emph{scenes}. Each scene carries an intensity that never drops from one
scene to the next, so the level ramps up.

\emph{(3) Translate to waves.} A deterministic step turns each scene's
enemy groups into concrete game waves. It reads the scene's free-text
intent to pick a formation and a movement---``flanking'' becomes a
pincer, ``V wedge'' a v-formation---and spreads each group's
\texttt{count} out in space. Without this step every wave would collapse
into a single column falling straight down.

\emph{(4) Stage trait content.} A second deterministic step (the
\emph{rule layer}) adds content tuned to the profile on top of the
draft. It adds three things. First, rewards that drop regardless of
kills: missile refills, plus---for players read as high collectors---
pickups placed in riskier spots. These give the Collector trait a way to
show itself that killing enemies cannot; without them a level might
never offer a pickup to decline in the first place (the
opportunity-scarcity problem of Section~\ref{app:records}). Second,
evenly spaced ability opportunities on the $\sim$10\,s berserk cooldown,
keyed to the Skill estimate, so how often the player spends an ability
is countable. Third, a set of numeric difficulty multipliers.

The full director prompt is in the Code and Data Supplement. The
director never sees $\theta$ or the raw features---only the derived
profile and the intensity number---so the ground-truth label cannot leak
through the generator.

\subsection{Content Style-Layer Templates}
\label{app:gen-templates}
The rule layer gives each of the three opportunity-mediated traits its
own content template, controlling reward placement, ability-opportunity
cadence, and enemy approach angles. Each template is built only from the
same wave settings a normal level uses---enemy HP multiplier
(\texttt{hpMul}$\le 2$), fire-interval multiplier
(\texttt{fireIntervalMul}$\ge 0.5$), enemy counts, and spawn cadence---
never from special measurement-only controls. A session played over a
template therefore serializes to the same leakage-free transcript as any
other level.
\begin{itemize}
\item \textbf{Collector (reward placement).} A standing center bullet
  field (a single \texttt{strong} enemy at \texttt{hpMul}$=2.0$, the
  schema cap, respawning every $2.6$\,s across $x\in\{140,200,260,320\}$
  with \texttt{fireIntervalMul}$=\mathrm{lerp}(1.4,0.6,c)$ so field
  density rises with the Collector estimate $c$), plus orb offers every
  $2.5$\,s whose $x$ is drawn from a safe-edge$\to$center ladder
  $[80,400,120,360,160,320,200,280,240]$. The estimate slides the
  offer-mix center ($\mathrm{lerp}(2,8,c)$ over the ladder): low
  collectors are offered mostly safe-edge pickups, high collectors
  mostly center (danger-priced) pickups, with a few of each kept as
  anchors.
\item \textbf{Aggression (enemy approach angles).} Sparse single tanky
  spawns (\texttt{strong}, \texttt{hpMul}$=1.6$, forward fire) on a
  $5$\,s cadence, entering from an alternating flank cycle
  $x\in[90,390,180,300,240]$; each is an independent pursue-or-ignore
  offer with a straight-down escape route, so pushing forward to engage
  is a choice the record captures rather than a forced encounter.
\item \textbf{Skill (ability-opportunity cadence).} A $12$\,s countable
  schedule that alternates a soft cluster ($8\times$ \texttt{diji1},
  berserk/missile bait) with a pressure spike ($3\times$ \texttt{strong}
  in a v-formation, \texttt{fireIntervalMul}$=0.6$, panic bait), evenly
  spaced so spend-per-opportunity is directly readable from the
  transcript.
\end{itemize}
Dodge is not templated: it is read from bullet-proximity behavior on
whatever field the other templates already produce, so it needs no
opportunity scaffold of its own.

\subsection{Closed-Loop Difficulty Refinement}
\label{app:gen-loop}
The mapping from a profile to its enemy-fire settings is deterministic:
the same profile yields the same settings and the same target on every
generation. What the refinement loop below adjusts is the assembled
\emph{content}, not that mapping, and it is applied only to the
synthetic downstream levels; the human study uses the draft directly.

Drafting a level does not pin how hard it actually plays: the director's
sense of ``how many enemies make intensity~7'' is noisy, so two levels
drafted for the same target can play far apart. To pin difficulty for
the synthetic downstream experiments, a refinement loop measures each
level and edits it. It plays the level with one \emph{fixed} bot---the
same bot for every level and target, so any difference in the reading
comes from the content and not from a changing player---and reads back
the difficulty the level actually produced, written $D_L^{\text{real}}$
(the difficulty coefficient $D_L$ of Section~\ref{app:difficulty},
measured on that play). It then edits the content until
$D_L^{\text{real}}$ meets the target (Algorithm~\ref{alg:loop}). The
bot's bias is calibrated once on the standard level
($D_L^{\text{real}}{=}0.573$, target${}=D^{*}{+}0.073$).

\begin{algorithm}[t]
\caption{Closed-loop difficulty refinement}
\label{alg:loop}
\begin{algorithmic}[1]
\REQUIRE level spec $s$, target $\tau$, tolerance $\epsilon{=}0.06$
\STATE density $\leftarrow 1$;\ \ escalation $\leftarrow 0$
\FOR{$i = 0$ \TO $\text{MAXITERS}$}
  \STATE $D \leftarrow$ probe-bot play of $s$; measure $D_L^{\text{real}}$
  \IF{$|D-\tau|\le\epsilon$} \RETURN $s$ \ENDIF
  \STATE adjust fire-density knob toward $\tau$ (proportional, on the logit scale)
  \IF{knob saturated \AND $D\ll\tau$}
    \STATE escalate enemy kinds: promote fodder waves (tail-first) to elite kinds (tracker fire $+$ hover dwell $+$ hp)
  \ELSIF{$D\gg\tau$}
    \STATE de-escalate one level
  \ENDIF
\ENDFOR
\end{algorithmic}
\end{algorithm}

The lever that actually moves difficulty is upgrading enemy \emph{type},
not adding more enemies. Turning up fire rate and wave count both stall
at $D_L^{\text{real}}\!\approx\!0.27$: weak, fast-leaving enemies
(\emph{fodder}) and undirected fire rarely come near the ship. Upgrading
a wave to a stronger \emph{elite} type changes three things at once---
the enemy aims its fire at the ship (\emph{tracker fire}), lingers on
screen instead of flying past (\emph{hover dwell}), and takes more hits
to kill. The single heaviest term in $D_L$, the share of aimed-fire
enemies (\texttt{trackerShare}), is zero on an all-fodder level, which
is why type is the effective knob. To keep the level's style, upgrades
are applied from the last scenes backward (preserving the ramp) and
reuse whatever elite type the level already uses. Across target levels
this cuts the mean absolute target error from $0.204$ (draft only) to
$0.011$, usually within $2$--$3$ plays.

\section{Dataset}
\label{app:dataset}

The Code and Data Supplement includes the synthetic benchmark: two
frozen $300$-session populations, each session
sampled at an independent random parameter vector under a fixed seed.
Every session ships with its ground-truth parameter label
(\texttt{oracleTraits}), the leakage-free transcript the reader
consumes, and a session summary (level, seed, outcome, duration). A
\texttt{schema.md} documents every field. Because a human's latent
traits are unobservable, no existing public dataset offers
verifiable trait labels; this is the gap the synthetic benchmark
fills, and the reason recovery accuracy can be measured at all.

\section{Human Study}
\label{app:human}

\subsection{Procedure}
Twelve participants each play a fixed baseline (calibration) level and
complete a short self-report questionnaire. Each then plays two further
personalized levels, in a counterbalanced order they are not told.
Both personalized levels are produced by the same LLM generator (the
director) from a four-axis player profile; the two conditions differ
only in where that profile comes from. In the \textbf{behavior-driven}
condition, the profile is the one the LLM infers from the participant's
own play. In the \textbf{questionnaire-driven} condition, the profile is
read straight off the participant's questionnaire self-report (Part~A)
and fed to the same generator in place of the inferred one. Because the
generator is held fixed, the comparison isolates the source of the
profile---read from behavior versus asked directly. After the sessions,
the LLM player model reads the behavioral records, and the ability
reading on each personalized level is compared against the baseline
reading.

\subsection{Questionnaire}
The instrument has two parts. Part~A is a self-report of play style: its
four axis scores form the profile that drives the questionnaire-driven
level, and are also compared against the LLM reading. Part~B collects
background. Responses are on a $7$-point Likert scale
($1={}$strongly disagree, $\dots$, $7={}$strongly agree) unless an item
says otherwise. Part~A is filled immediately after the baseline level
and \emph{before} the participant sees any model output, to prevent
anchoring; Part~B is filled last.

\paragraph{Part A --- self-reported play style.} Each axis pairs a
forward ($+$) and a reverse ($-$) item; the axis score is
$\mathrm{mean}(\text{forward},\,8-\text{reverse})$, normalized to
$[0,1]$ by $(\text{score}-1)/6$. These four scores are the profile
passed to the generator in the questionnaire-driven condition.
\begin{itemize}\setlength\itemsep{1pt}
\item[A1] \emph{(Dodge, $+$)} Even when bullets are dense, I can steadily avoid most of them.
\item[A2] \emph{(Dodge, $-$)} When I get hit, it is mostly because I could not react in time.
\item[A3] \emph{(Collector, $+$)} Whenever a pickup or supply appears, I try to go get it.
\item[A4] \emph{(Collector, $-$)} If grabbing a pickup is risky, I would rather skip it.
\item[A5] \emph{(Aggression, $+$)} I tend to push to the front of the screen to attack rather than stay in a safe zone.
\item[A6] \emph{(Aggression, $-$)} I prioritize survival first, and only then think about kills.
\item[A7] \emph{(Skill, $+$)} As soon as an ability (missile / berserk) is ready, I tend to spend it rather than hoard it.
\item[A8] \emph{(Skill, $-$)} I often finish a run with abilities still unused.
\end{itemize}

\paragraph{Part B --- background.}
\begin{itemize}\setlength\itemsep{1pt}
\item[B1] Experience with bullet-hell / flight-shooter games: $1={}$never played, $\dots$, $7={}$very experienced.
\item[B2] Average weekly time spent playing games: none / ${<}2$h / $2$--$7$h / $7$--$15$h / ${>}15$h.
\end{itemize}

\subsection{Analysis}
The primary readout plotted in the study figure is the change in the
macro ability reading $\hat\theta$ (the mean over the four axes) from
the baseline level to each personalized level; the comparison of
interest is whether the behavior-driven level lifts $\hat\theta$ more
than the questionnaire-driven level. As a secondary check, the
per-axis Part~A self-report is compared with the LLM reading, with the
self-report treated as a noisy, potentially circular criterion rather
than as accuracy. At $n{=}12$ these are reported as directional, not
significance-tested.

\end{document}